\documentclass[conference]{IEEEtran}
\IEEEoverridecommandlockouts
\usepackage{cite}
\usepackage{amsmath,amssymb,amsfonts}
\usepackage{algorithmic}
\usepackage{graphicx}
\usepackage{textcomp}
\usepackage{xcolor}
\usepackage{graphicx}
\usepackage{bm}
\usepackage{amssymb}
\usepackage{balance}  
\usepackage[caption=false]{subfig}

\def\BibTeX{{\rm B\kern-.05em{\sc i\kern-.025em b}\kern-.08em
    T\kern-.1667em\lower.7ex\hbox{E}\kern-.125emX}}
\begin{document}

\title{Online Changepoint Detection on a Budget \\

}

\author{\IEEEauthorblockN{Zhaohui Wang}
\IEEEauthorblockA{\textit{Applied Research} \\
\textit{Splunk}\\
San Francisco, US\\
zhaohuiw@splunk.com}
\and
\IEEEauthorblockN{Xiao Lin}
\IEEEauthorblockA{\textit{Applied Research} \\
\textit{Splunk}\\
San Francisco, US\\
xlin@splunk.com}
\and
\IEEEauthorblockN{Abhinav Mishra}
\IEEEauthorblockA{\textit{Applied Research} \\
\textit{Splunk}\\
San Francisco, US \\
amishra@splunk.com}
\and
\IEEEauthorblockN{Ram Sriharsha}
\IEEEauthorblockA{\textit{Applied Research} \\
\textit{Splunk}\\
San Francisco, US \\
rsriharsha@splunk.com}
}

\maketitle

\begin{abstract}
Changepoints are abrupt variations in the underlying distribution of data. Detecting changes in a data stream is an important problem with many applications. In this paper, we are interested in changepoint detection algorithm which operates in an online setting in the sense that both its storage requirements and worst-case computational complexity per observation are independent of the number of previous observations. We propose an online changepoint detection algorithm for both univariate and multivariate data which compares favorably with offline changepoint detection algorithms while also operating in a strictly more constrained computational model. In addition, we present a simple online hyperparameter auto tuning technique for these algorithms. 
\end{abstract}

\begin{IEEEkeywords}
changepoint, streaming, online, detection
\end{IEEEkeywords}

\section{Introduction}

Today, unbounded data is streamed in unprecedented volumes and varieties, in diverse domains such as application logs and metrics monitoring, wearable devices, sensor devices, and more. 
In many of these domains, the ability to analyze data on the stream is valuable from an early detection and response perspective, providing interesting challenges and opportunities for algorithm designers\cite{disabato2021tiny, gama2014survey}.

When analyzing data offline, it is reasonable to assume the data was generated by a fixed process; for example, the data is a sample from a static (albeit multimodal) distribution. 
However, on the stream, there is a temporal dimension, and the generative parameters of a data stream can change.
The quantification and detection of changepoints is one of the fundamental challenges in the streaming setting. Batch machine learning techniques trained on previous datasets need to undergo parameter update through retraining.

The field of changepoint detection has a long history (see \cite{aminikhanghahi2017survey} for an overview).
However, the vast majority of this growing body of literature focused on the retrospective segmentation problem (see \cite{truong2020selective} for an overview), where, after the entire data stream is observed, the algorithm has to detect any changepoints.

In the streaming model of computation, any changepoint detection algorithm must consume the data in one pass and is allowed to keep only a small (typically constant or poly-logarithmic in n which is the number of data points) amount of information.

The online model of computation does not allow us to indefinitely postpone when to output a changepoint. In this setting, a priori unknown number of points arrive one by one in an arbitrary order. 
When a new point arrives the algorithm must either flag this point as a changepoint or decide whether the generative parameters of the data distribution have sufficiently drifted for this point to be considered a changepoint. 
The quality of such an algorithm depends on how much data one needs to see to determine a distribution shift. At the same time, subsequent delay in detecting changepoints cause staleness in trained models. The faster a changepoint is detected, the faster we can update a machine learning model.

In this paper, we consider the intersection of these two models of computation. We designed online changepoint detection algorithms on a budget which store constant amount of information and are independent of the size of the stream. We summarize the contributions as follows:
\begin{itemize}
  \item Online changepoint detection algorithms that work on unbounded data stream with a constant time and space complexity. 
  \item Along with univariate drift detection, multivariate case is studied where covariance drift occurs.
  \item A simple hyperparameter auto tune approach is proposed to quickly warm up the online algorithm.
\end{itemize}

\section{Related Work}
The first (to our knowledge) result in online changepoint detection dates back to Page \cite{Page1954, Page1955} and Lorden \cite{lorden1971}. 
Here it is assumed the parameters of the distribution are known and one performs a sequential likelihood test to determine if the current point represents a changepoint in the mean of these parameters. 

A generalization of this approach is to analyze the probability distributions of data before and after a candidate changepoint, and identify the candidate as a changepoint if the two distributions are significantly different \cite{kifer2004detecting, Tartakovsky2006DetectionOI}. Such approaches are in general sensitive to the choice of window sizes, thresholds and of course the distance function that measures the distance between the two distributions.

Deep learning and non-parametric approaches are introduced into drift detection \cite{gama2014survey, hinder2020towards, 9513865}, which solve for non-parametric drift detection of diverse drift characteristics. And enable to rely on independence tests rather than parametric models or the classification loss. However, deep learning based approaches indicate higher requirements for computing power and infrastructure.

Recently, a Bayesian perspective on univariate online changepoint detection was provided by Adams and McKay \cite{Adams:aa}, where the model parameters before and after the
changepoint are examined, and therefore the probability distribution of the length of the
current run is computed. This construction is not a streaming algorithm, in the sense that the worst case storage requirement can be O(n).

We will however present two straightforward modifications of Adams and MacKay's construction \cite{Adams:aa}, that taken together allows us to both detect multivariate changepoints online as well as satisfy the storage requirements of a streaming algorithm by maintaining a fixed storage independent of n. We will also show how the hyperparameters of the algorithm can be auto-tuned.

\section{Methodology}

We study the Bayesian online changepoint detection algorithm of \cite{Adams:aa} and its storage bounds.
This will suggest natural extensions to the streaming as well as multivariate settings.

\subsection{Notation and Preliminaries}
We assume a stream of observations $\bm{x}_1$, $\bm{x}_2$, ..., $\bm{x}_T$ may be divided into non overlapping partitions, or changepoints. 
Within each partition $\rho$, the data points are drawn i.i.d. from some underlining probability distribution $P(\bm{x}_i|\bm{\eta}_\rho)$.
The parameters $\bm{\eta}_\rho$ are taken to be i.i.d as well.
However, between each partition, the underlining data distributions can be different.
We denote the contiguous set of observations between time $a$ and $b$ inclusive as $\bm{x}_{a:b}$ and the time since the last changepoint as $r_t$, the run length.
We also use the notation $\bm{x}_t^r$ for the set of observations associated with $r_t$.

\begin{figure}%
    \centering
    \subfloat[]{{\includegraphics[width=6cm]{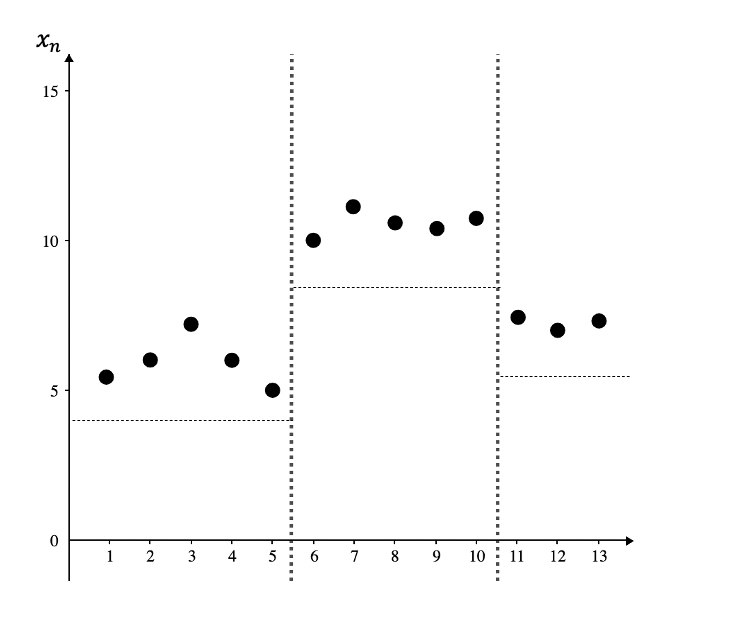} }}%
    \qquad
    \subfloat[]{{\includegraphics[width=6cm]{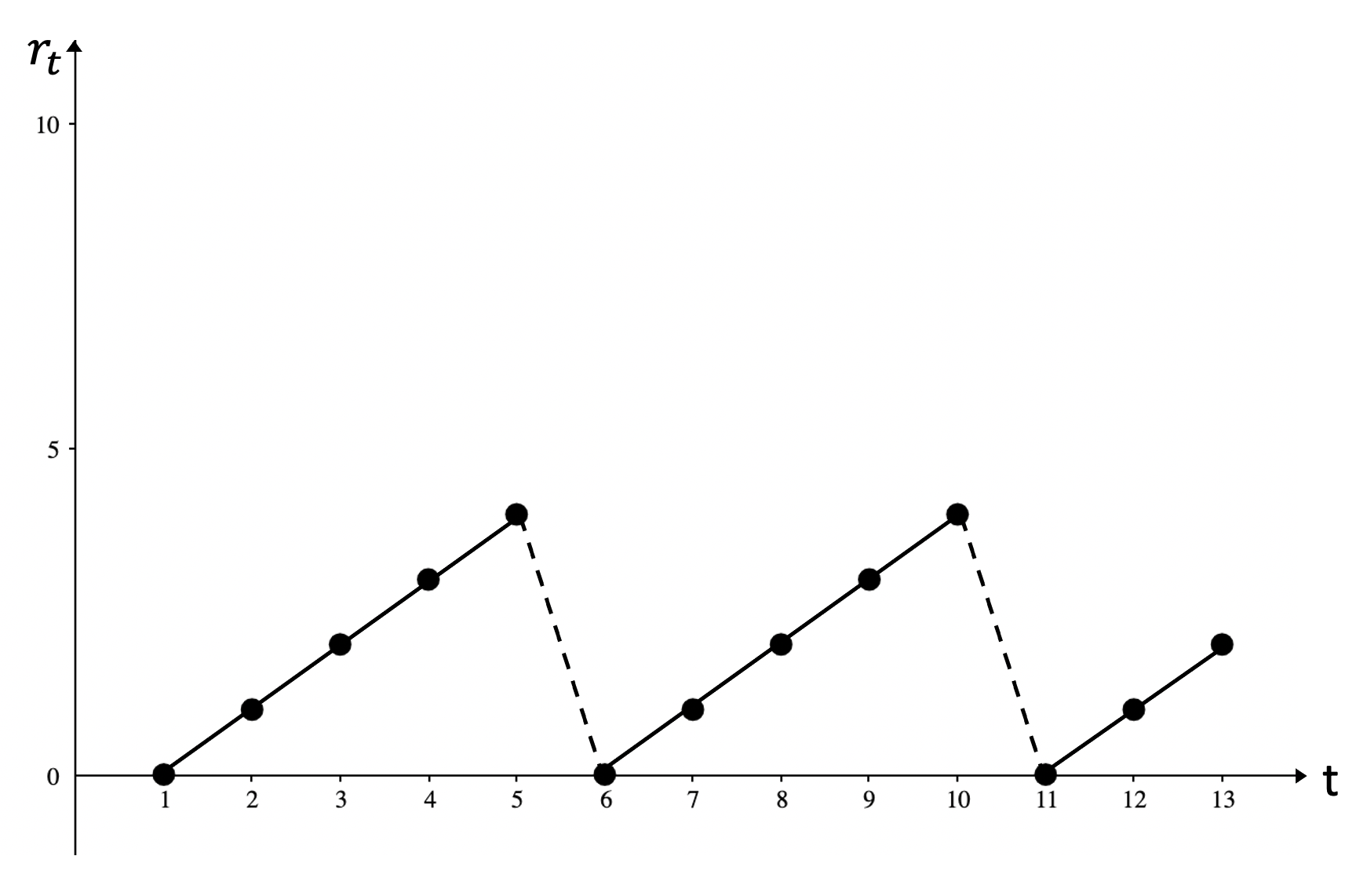} }}%
\caption{An example for a sequence of data samples (a) and corresponding run length 5, 5 and 3 (b).}
\label{illus1}
\end{figure}

Figure \ref{illus1} illustrates the relationship between run lengths and parameters of univariate data.

\subsection{Bayesian Online Changepoint Detection}

Following \cite{Adams:aa}, the idea is to estimate the posterior distribution over the current run length $r_t$ given data points we have seen so far.

The posterior distribution of the current run length  $P(r_t|\bm{x}_{1:t})$ is computed as:
\begin{equation}
  P(r_t|\bm{x}_{1:t}) = \frac{P(r_t, \bm{x}_{1:t})} {P(\bm{x}_{1:t})}
\end{equation}

We write the joint distribution over the run length and the observed data in the form:
\begin{equation}\label{eq:eqn1}
  \renewcommand\arraystretch{1.25}
    \begin{aligned}[b]
      P(r_t, \bm{x}_{1:t}) &= \sum_{r_{t-1}}P(r_t, r_{t-1}, \bm{x}_{1:t})  \\
        &=  \sum_{r_{t-1}}P(r_t, x_t|r_{t-1}, \bm{x}_{1:t-1}) P(r_{t-1}, \bm{x}_{1:t-1}) \\
        &= \sum_{r_{t-1}}P(r_t|r_{t-1})P(x_t|r_{t-1}, \bm{x}_{t}^r) P(r_{t-1}, \bm{x}_{1:t-1})
    \end{aligned}
  \end{equation}

By integrating over the posterior distribution on the current run length, we obtain the marginal predictive distribution.
\begin{equation}
  P(\bm{x}_{t+1}| \bm{x}_{1:t}) = \sum_{r_t} P(\bm{x}_{t+1}|r_t, \bm{x}_t^r)P(r_t|\bm{x}_{1:t})
\end{equation}
 
The current run length at time t only has two possible outcomes at the next time point $t+1$.
\begin{equation}\label{prior1}
 r_{t+1} = \begin{cases}
  0 &\text{     if } \bm{x}_t \text{ is  a changepoint}\\
  r_t + 1 &\text{     }  \text{otherwise}
  \end{cases}
\end{equation}

This suggests a computationally efficient conditional prior on the changepoint in terms of the \textit{hazard function}. 
\begin{equation}\label{prior2}
  P(r_t|r_{t-1})= \begin{cases}
   H(r_{t-1} + 1) &\text{if } r_t = 0\\
   1 - H(r_{t-1} + 1) &\text{if } r_t = r_{t-1} +1\\
   0 &\text{otherwise}
   \end{cases}
 \end{equation}

A special case, well suited to streaming applications, is where the conditional prior is memory-less, leading to a constant hazard function $H(\tau) = 1/ \lambda$.

While the algorithm of \cite{Adams:aa} is online, it requires us to store all run length probability estimates up to the current data point in order to estimate $P(r_t, \bm{x}_{1:t})$. Therefore, its space and time complexity for each new data point is O(n).
A trivial modification, suggested in \cite{Adams:aa} is to simply discard the run length probability estimates in the tail of the distribution which have a total mass less than some threshold.
While this improves the average case complexity per new data point, it can still lead to a worst-case complexity of O(n).

\subsection{Bounding the Storage}
A natural extension  is to maintain a fixed size buffer of size $L$.
We keep a record of the starting index of each run denoted as $[i_{r_0}, i_{r_1}, . . . , i_{r_L}]$. When a data point $\bm{x}_j$ arrives, the current starting indices become $[i_{r_0}, i_{r_1},..., i_{r_n}, j]$ with length $L+1$. By calculating the posterior run length probability at this point, we evict the index of with lowest probability, say $k$. Then, the starting indices become $[i_{r_0}, i_{r_1},...,i_{r_{k-1}},i_{r_{k+1}}, ..., i_{r_n}]$ with length $L$ again.

In this fashion, we drop the run length starting at the index with lowest posterior probability while keeping the memory usage fixed. We now have a few options to distribute the evicted probability mass. Empirically, we also found that the simplest approach of dropping the probability mass and not redistributing it works well on a variety of datasets.

\subsection{Multivariate Changepoint Detection}
For data streams originating from the normal distribution $(\mu, \tau)$, we propose non-informative conjugate prior normal-gamma with parameters $(\mu, \kappa, \alpha, \beta)$. 
In univariate case, we obtain a closed form posterior Student T distribution. For each $x_i$, we update the parameters as follows:
\begin{align*}
  \alpha &= \alpha + 0.5 \\
  \beta &= \beta + \frac{\kappa*(x_i-\mu)^2}{2*(\kappa+1)}\\
  \mu &= \frac{(\kappa * \mu + x_i)}{\kappa + 1}\\
  \kappa &= \kappa + 1
\end{align*}

The most common approach in multivariate drift detection is to detect drift on each individual univariate, which ignores the non-stationary case when covariance drift happens.
To properly detect drift for all the non-stationary cases,  we  specify a multivariate normal distribution and normal inverse Wishart prior \cite{wishart1928generalised}, and thus obtain a closed form posterior. 
However, we observed that excessive false positives are detected. To improve, and to conserve the same update rules as the univariate case, and therefore ensuring the algorithm is robust to the data dimension,
we propose a posterior Student T distribution where the parameters updates are similar to the univariate case. For each $\boldsymbol{x_i}$,
\begin{align*}
  \alpha &= \alpha + 0.5 \\
  \boldsymbol{\beta} &= \boldsymbol{\beta}+ \frac{\kappa*((\boldsymbol{x_i}-\boldsymbol{\mu)} * (\boldsymbol{x_i}-\boldsymbol{\mu}) ^ T)}{2*(\kappa+1)}\\
  \boldsymbol{\mu} &= \frac{(\kappa * \boldsymbol{\mu} + \boldsymbol{x_i})}{\kappa + 1}\\
  \kappa &= \kappa + 1
\end{align*}
With these update rules, we have two benefits. First, we take multivariate covariance structure into the detection. Second, we obtain the exact same update rule as univariate case without any loss of generality.

\subsection{Tuning Hyperparameters Online}
Through testing on a variety of real world datasets, we observed the algorithm of \cite{Adams:aa} is quite sensitive to initial hyperparameter settings. The details of which are discussed further in the experimentation section. For example., excess false positives in particular may arise from the cold start of the algorithm, where the hyperparameters fail to adapt to the data distribution quickly.

To speed up the learning process and automatically adapt the algorithm to incoming data in an online fashion, we propose an auto hyperparameter tuning approach. 
Since we are inferring the posterior Student T distribution of run length, we  estimate its mean $\mu_x$ and variance $\sigma_x^2$ with a sample of data observed so far. Therefore we reduce two degrees of freedom for the initial values of $\beta$ and $\mu$, we denote the initial value for $\beta$ as $\beta_0$, and correspondingly for the other hyperparameters,
\begin{align*}
   \beta_0 &= \frac{\alpha_0 * \kappa_0 * \sigma_x^2}{\kappa_0 + 1}\\
   \mu_0 &= \mu_x
\end{align*}

The initial values for both $\beta$ and $\mu$ are estimated from the data observed from the stream. In this way, we  warm up the algorithm quickly and efficiently, where hyperparameters are auto adapted to the data observed instead of being hard-coded to initial settings.
For efficiency, we simply put the first several data points in the buffer to estimate the hyperparameters, i.e., mean and variance of the first several data points are calculated and used as estimated mean and variance for the posterior distribution.
We observe that the size of this buffer has little effect on the warm up process, as long as a broad estimation of the distribution parameters can be inferred.

\begin{table*}
\centering
\caption{Univariate model performance comparison (MAE). Results in bold represent the lowest MAE.}
\begin{tabular}{|c|cccccc|} \hline
Algorithm&Martingale& Temp.&Nile&Earthquake & normal & normal and Uniform\\ \hline
\multicolumn{7}{|c|}{Offline Algorithms }  \\ \hline
BinSeg & 1,004 & 600 & 100 & 9,000 & NA & NA\\ 
PELT & \textbf{5} & 3,044 & 107 & 5,978 & \textbf{102} & NA\\ 
SegNeigh  & \textbf{5} & NA & 107 & 3256 & NA & NA\\ 
Exch. Martingale & 851 & 67& 100 & 55,000,000&NA&NA \\ 
KS&185,811  & 131,200 & 304 & 456,009 & 2,853,287,032 &2,763,991,675\\ \hline
\multicolumn{7}{|c|}{Online Algorithms}  \\ \hline
EWMA& 17 & 1,376 & \textbf{5} & 5,296 & 1,849,476 & 13,256\\ 
AFF  &  26 & 1,376 & 8  &  5,301 & 1,139,428 & 6986\\ 
FFF & 141 & \textbf{0} &163 & 5,972 & 6,831&10,791 \\ 
BCPD & 2,430 & 83 & 100 & 104,447 & NA & NA\\ \hline
\multicolumn{7}{|c|}{Online Streaming Algorithms}  \\ \hline
CUSUM & 38 & 688 & \textbf{5} & 5,296& 35,561 & 21,903\\ 
Online BCPD &988 & 45 & 23 & \textbf{3,161} & \textbf{0} & \textbf{0}\\ \hline
\end{tabular}
\label{uni_performance}
\end{table*}

\begin{table*}
       \centering
       \caption{Multivariate model performance comparison (MAE)}
       \begin{tabular}{|c|ccccc|} \hline
       Algorithm& Mean change&Variance change&Covariance change&Sentiment data&Concept drift\\ \hline
       R ECP& 1  & 1 & 100 & 19& NA\\ 
       KL Divergence & 2 &15 & 47 & 138 & 7,200\\ 
       Offline BCPD& \textbf{0}  & \textbf{0}  & 100 & 34& 15\\ 
       Online BCPD& \textbf{0} &\textbf{0} & \textbf{34} & \textbf{12} & \textbf{0}\\ \hline
       \end{tabular}
       \label{mul_performance}
\end{table*}

\section{Experimentation}
In this section, we evaluate model performance of both univariate and multivariate drift detection on both real world and synthetic datasets. Furthermore, we compare the online BCPD algorithm to state-of-the-art drift detection algorithms.
The test datasets consist of Martingale \cite{ho2007detecting}, global historical temperature \cite{globaltemp}, Nile \cite{nile} and earthquake data from Northern California Earthquake Catalog \cite{xie2019asynchronous}. Aside from real world data, we also generate datasets where changepoints are known. We construct a
stream of 1 million points and change the normal distribution every 10,000 points, as well as another dataset of the same size where we vary the distribution between normal and uniform distributions.

The benchmark algorithms we tested are binary segmentation (BinSeg) \cite{scott1974cluster, rcp}, PELT \cite{killick2012optimal}, segment neighbourhoods (SegNeigh) \cite{auger1989algorithms}, exchangeable with martingales (Exch. Martingale) \cite{ho2007detecting}, Kolmogorov-Smirnov (KS) statistic with a window size 20, 
as well as the original Bayesian changepoint detection (BCPD) \cite{Adams:aa}. We also compare with online algorithms such as AFF \cite{bodenham2017continuous}, FFF \cite{bodenham2017continuous}, CUSUM \cite{Page1954} and EWMA \cite{roberts2000control}. 
\subsection {Metrics}
Many metrics are used to measure changepoint detection performance \cite{aminikhanghahi2017survey}. However, we find some metrics fail to take false positive and false negative into consideration.  
To properly compare accuracy of different algorithms, we design a mean absolute error (MAE) with penalty as our metric. For any actual changepoints
$actual = [\gamma_1, \gamma_2, ...,  \gamma_j]$ and predicted changepoints $ predicted = [p_1, p_2,..., p_k]$ with a total number of n data points, the MAE loss is defined as
\begin{equation}\label{mae}
  loss = \sum_i|\gamma_i - p_i| + penalty
 \end{equation}
where the penalty for false negative or false positive is,
\begin{equation}\label{mae2}
  penalty = \begin{cases}
   (j-k)n &\text{if } j \geq k\\
   \sum_k|p_k| &\text{if } j < k
   \end{cases}
 \end{equation}
The loss is calculated between the closest point pairs $\gamma_i$ and $p_i$. In the case of false negatives, we penalize the number of data points that are missed by the detector, which is the data length n times the number of missed positives. 
While in the case of false positive, we penalize the excessive detected points.
\subsection{Univariate Performance}
Univariate drift detection performance is shown in Table \ref{uni_performance} where NA values indicate the corresponding model cannot run due to data scale being too large to handle. We break down the table into 3 parts, offline, online and online streaming algorithms. From 3 of the 7 datasets, our online streaming drift detection algorithm presents the best MAE (Mean Absolute Error).It worth to note KS outputs too many false positives. To apply KS properly, we need to provide a reasonable window size which is difficult in the online setting. Moreover, with automatic hyperparameter tuning, we see online streaming BCPD outperforms original BCPD.

\begin{figure}
  \includegraphics[width=1\linewidth]{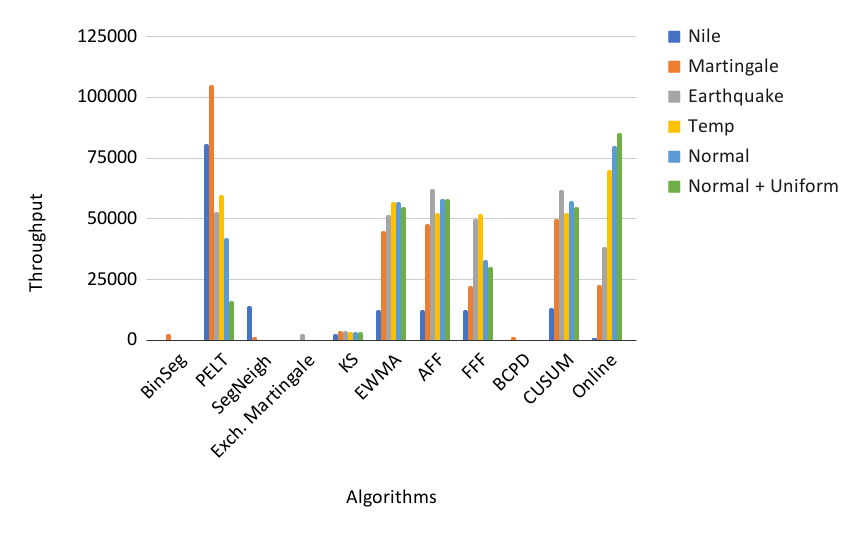}
  \caption{Throughput comparison, the datasets are sorted from small size to large size (from left: 100, 4000, 14998, 100000, 1000000, 1000000 records correspondingly). Online algorithms present top throughput compared to offline algorithms. Online BCPD demonstrates comparable throughput among online algorithms.}
  \label{fig:thr1}
\end{figure}

More importantly, throughput comparison reported in Figure \ref{fig:thr1} demonstrate the much improved efficiency of our streaming algorithm. The datasets are sorted from smallest to largest. We observe the larger the dataset, the more processing speed advantage streaming drift detection brings.

\begin{figure}
  \includegraphics[width=\linewidth]{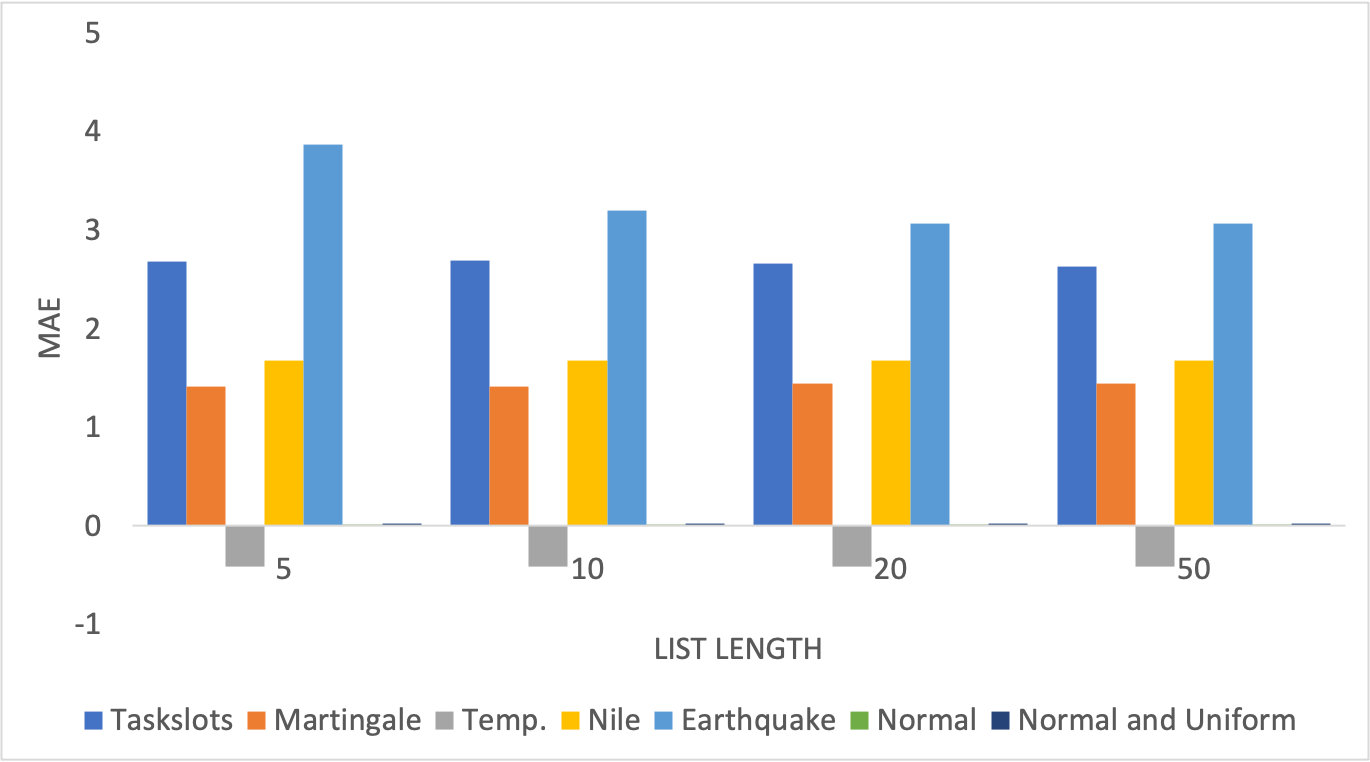}
  \caption{Online BCPD performance is robust to the choice of budget size for memory usage. MAE error is reported in log scale.}
  \label{fig:robust1}
\end{figure}

We also emphasize our approach is robust to the length of list budget. As in Figure \ref{fig:robust1}, with a minimum of list length 10, we can obtain stable model performance and score millions of data online.

\subsection{Multivariate Performance}
Our online streaming multivariate drift detection algorithm has two built-in properties. First, it preserves exactly the same univariate drift detection result if taking variables one by one, without any loss of generality. 
Second, it is able to properly detect covariance/correlation drift. We benchmark multivariate changepoint detection with non parametric multivariate changepoint detection \cite{ecp_paper, ecp_package} and KLL \cite{kullback1951information}.

To test the algorithm, we simulate datasets with multivariate normal distribution, where mean, variance and covariance drifts are generated as follows: 

For the mean case, we simulate 2 dimensional gaussian ($\mu$ = [1, 0], $\sigma$ = [1, 0, 0, 1]) and ($\mu$ = [10, 0], $\sigma$ = [1, 0, 0, 1]). 

For the variance drift, we simulate gaussian ($\mu$ = [1, 0], $\sigma$ = [1, 0, 0, 1]) and ($\mu$ = [1, 0], $\sigma$ = [1, 0, 0, 10]). 

\begin{figure}
    \minipage{0.2\textwidth}
      \includegraphics[width=\linewidth]{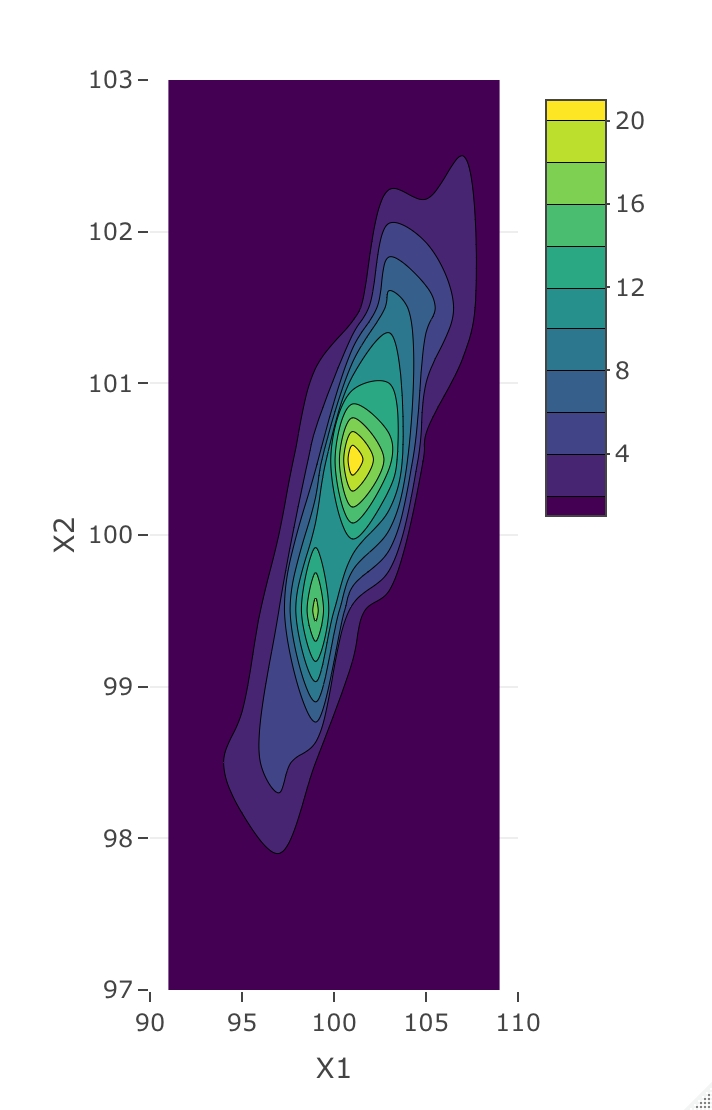}
    \endminipage\hfill
    \minipage{0.2\textwidth}
      \includegraphics[width=\linewidth]{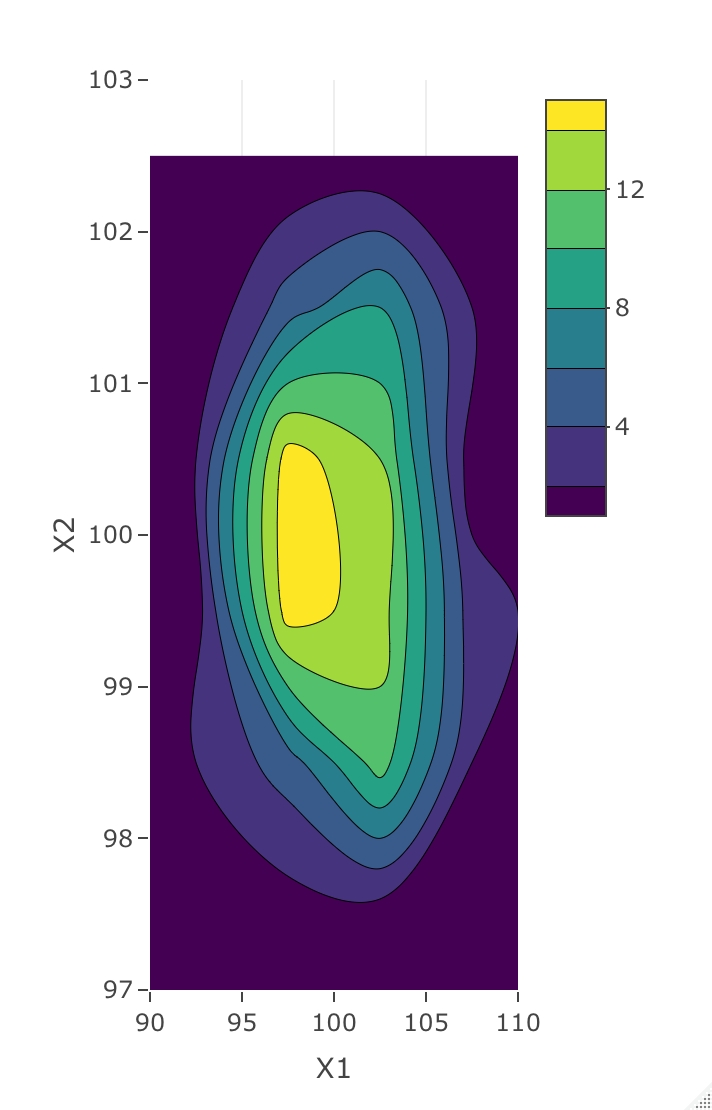}
    \endminipage
\caption{An example of bi-variate normal samples with correlation 0.3 (left) and 0 (right).}
\label{fig:cov3}
\end{figure}

For the covariance drift, gaussian ($\mu$ = [1, 0], $\sigma$ = [1, 0, 0, 1]) and ($\mu$ = [1, 0], $\sigma$ = [1, 0.3, 0.3, 1]) is simulated as in Figure \ref{fig:cov3}. 

Additionally, synthetic concept drift data are generated (two dimensional circular function before drift and eclipse function after drift).
Moreover, we test on real world WalBelSentiment \cite{rpackage} dataset with sentiment changes.

As seen in Table \ref{mul_performance}, our algorithm provides the same accuracy as the offline version in detecting mean and variance drift without any loss.
However, the offline algorithm fails to detect covariance change while our multivariate approach is sensitive to it . 
On the real multivariate sentiment drift dataset, the online streaming algorithm again outperforms competitors. 

\subsection{Hyperparameters Auto Tune}
\begin{figure}
{%
  \includegraphics[clip,width=8cm, height=5cm]{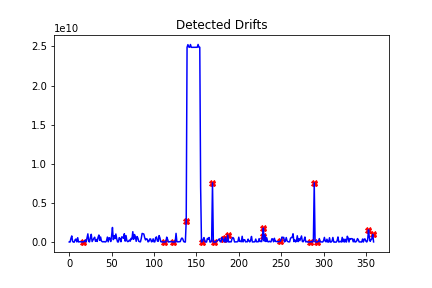}%
  
}
{%
  \includegraphics[clip,width=8cm, height=5cm]{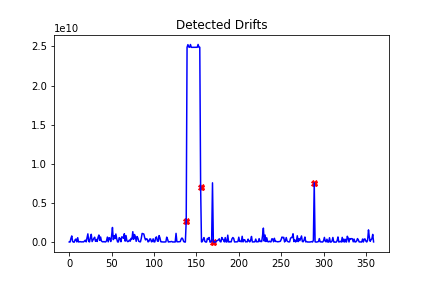}%
}
\caption{Online BCPD detected drifts with default hyperparameter (above) and with hyperparameter auto tune (below). With default hyperparameter, we observe 11 more false positives.}
\label{auto}
\end{figure}

\begin{figure}
  \includegraphics[width=\linewidth]{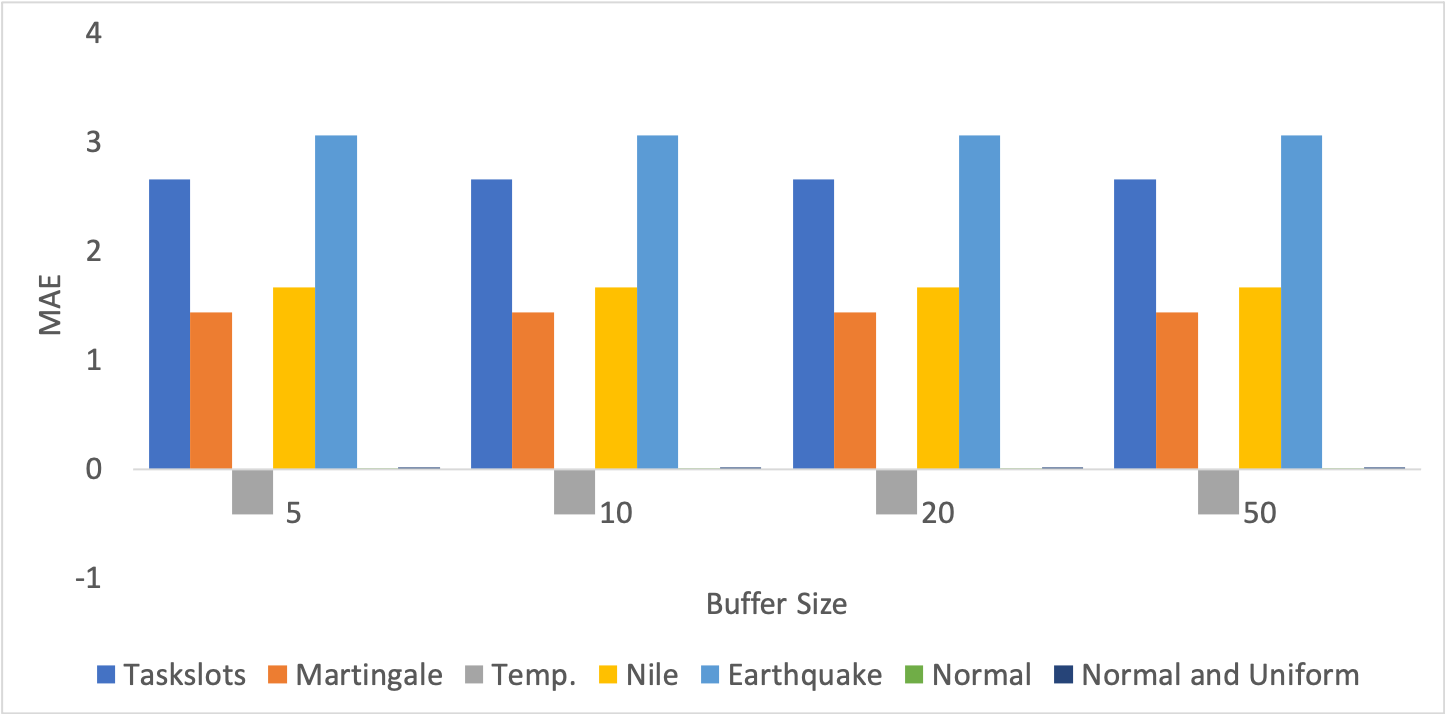}
  \caption{Online BCPD performance is robust to the buffer size for hyperparameter auto tune. MAE error is reported in log scale.}
  \label{fig:robust2}
\end{figure}

The most common problem of an online streaming algorithm faces is the volume of data. We not only require a high throughput, but also the ability to adapt to the stream quickly. We create a real-world dataset to conduct performance tests. The data is compiled from an enterprise PAN firewall 16 days history log that contains more that 148 million events. The compiled data events are ingested into Apache Flink cluster to emulate an unbound streaming environment where our algorithm can perform drift detection. The detection is paralleled based on unique key of each source IP address of firewall events. The data ingestion in emulation is keeping a high throughput (more than 30,000 events per second) so that ingestion will not become the bottleneck in this streaming analysis setup.  Through testing our algorithm on this real-world dataset, we found the model is quite sensitive to hyperparameter settings, shown in Figure \ref{auto}, which represents a single source IP’s network traffic and red crosses mark the detected change points. The default hyperparameter we set originally are: $ \lambda = 250$, $\alpha = 0.1$, $\beta = 0.01$, 
$\kappa = 1$ and $\mu=0$. With this default setting, we observed too many false positives when deploying online as shown in Table \ref{cps detected}. Therefore we propose an online tuning approach to automatically tune the hyperparameters based on the buffered observed data. As shown in Figure \ref{fig:robust2}, model performance is robust to the buffer size. 

\begin{table*}
  \centering
  \caption{Hyperparameter effects on number of changepoint detected}
  \begin{tabular}{|c|c|c|c|} \hline
  Data series & Data length & \multicolumn{2}{|c|}{Numbers of detected CPs }\\
  \hline
   & & Default hyperparameters & Auto tuning\\\hline
  Bytes Received by Hour& 360&15&3\\ 
  Bytes Received by Minute& 21,600&1,280&86\\
  Bytes Received by Second& 1,296,000&57,737&37,112\\ \hline
  \end{tabular}
  \label{cps detected}
  \end{table*}

\begin{table*}[]
\centering
\caption{Model performance comparison with non Gaussian distributions (MAE)}
\begin{tabular}{|c|ccccc|} \hline
Algorithm&Poisson&Gamma & Lognormal&Mixed Gaussian &Lognormal with Normalization\\ \hline
\multicolumn{6}{|c|}{Offline Algorithms }  \\ \hline
BinSeg &2&3&4,904&5,500 & 9,000\\ 
PELT &68,793&\textbf{0}&4,741&11,424& 5,619\\ 
SegNeigh &3,000&3,504&4,877&4,500 & 2,500\\ 
Exch. Martingale&890&1,060&\textbf{411}&670 & 1,470\\ 
KS&87,7447&3,959&89,245&69,098 & 90,,663\\ \hline
\multicolumn{6}{|c|}{Online Algorithms }  \\ \hline
EWMA& 88 &75&4,297&671 &51\\ 
AFF& 72 &54&4,297&70&72\\ 
FFF& 90 &63&4,362&54 &81\\ 
BCPD&\textbf{0} &\textbf{0}&4,690&\textbf{0} &\textbf{0}\\ \hline
\multicolumn{6}{|c|}{Online Streaming Algorithms }  \\ \hline
CUSUM& 187 &176&4,128&279 &125\\ 
Online Streaming BCPD&\textbf{0} &\textbf{0}&1,423&\textbf{0} & \textbf{0}\\ \hline
\end{tabular}
\label{nongau}
\end{table*}

\subsection{Non Gaussian Distributions}
Although gaussian distributed data stream is the most common use case, we may encounter non gaussian cases as well. In this section, we want to explore algorithm robustness to non gaussian distributions. We simulate 10,000 data points with 10 known changepoints, including poisson ($\lambda$ = 2.0 and $\lambda$ = 10.0), gamma ((shape = 2.0, scale = 2.0) and (shape = 10.0, scale = 10.0)), lognormal (($\mu$ = 3.0, $\sigma$ = 1.0) and ($\mu$ = 10.0, $\sigma$ = 1.0)) and mixed gaussian (a mixture of ([5,1], [1, 1.3], [9, 1.3]) and ([50, 1], [5, 1.3], [9, 5])).

Table \ref{nongau} shows the online streaming model performance remains stable for poisson, gamma and mixed gaussian distributions. Martingales provides the best accuracy for lognormal. BCPD gaussian algorithm is robust to data stream distributions, while a configurable distribution assumption is desired. Otherwise, normalization on the stream can serve as a preconditioner to the model. With normalization on the lognormal data, we see gaussian BCPD regains accuracy.

\subsection{A Comparison with CUSUM}
In this section, we specifically compare performance with online streaming CUSUM since it is also online and budget friendly. Performance is presented in the previous section. We point out that we tuned CUSUM with the best hyperparameters so it performs the best on our benchmark datasets. For the Martingale dataset, we know beforehand drifts happen each 1000 points. To obtain the best MAE, we manually set the burn in to 1000. Note how performance varies with different burn in as seen in Table \ref{cusum-mar}. However, tuning online is generally difficult for an online streaming algorithm, making CUSUM hard to configure. 

\begin{table*}[!htbp]
\centering
\caption{Sensitivity to hyperparameters on Martingale dataset (MAE)}
\begin{tabular}{|c|ccccc|} \hline
&burn in = 5& burn in = 20& burn in = 100& burn in = 500& burn in = 1000\\ \hline
CUSUM & 70,329 &6,575&25,041& 5,090 & 38 \\ \hline
\end{tabular}
\label{cusum-mar}
\end{table*}

Moreover, in real scenarios we are often facing extensive outliers in the data stream. We compare both algorithms with varies percentage of outliers inserted in the 1 million normally distributed dataset, and the results are listed in Table \ref{cusum}. It shows the online streaming BCPD algorithm significantly outperform against CUSUM method when outliers are presented.

\begin{table*}[!htbp]
\centering
\caption{Sensitivity to outliers (MAE)}
\begin{tabular}{|c|cccc|} \hline
&0.1 percent outliers& 0.5 percent outliers & 1 percent outliers& 10 percent outliers \\ \hline
CUSUM & 52 &57& 66 & 38 \\ 
Online Streaming BCPD&\textbf{0} &\textbf{0}&\textbf{0} & \textbf{0}\\ \hline
\end{tabular}
\label{cusum}
\end{table*}

\section{Conclusions}
We contribute online streaming Bayesian changepoint detection algorithms that work on unbounded data stream with a constant time and space complexity, for both univariate and multivariate cases.
The hyperparameter auto tune approach serves as a conditioner for the online algorithm to quickly warm up. The multivariate approach is able to successfully detect covariance drift.
More importantly, we have seen superior throughput with previous approaches, while providing comparable accuracy.

\section{Future Work}
We demonstrate the online streaming algorithm is robust to data distributions, but this can be further improved with normalization. Different data distribution assumptions and associated update rules can also be further expanded on.

Our multivariate approach relies on constructing an n by n matrix, which is expensive when dimension n of the feature space is large. A more efficient way to shrink matrix dimension in run probability calculation while preserving reasonable accuracy and throughput is desired.

Finally, this online streaming approach serves as an initial step for continual learning, especially for learning in the presence of drift. A connection between run length probability and optimizer learning rate would be an interesting topic to further investigate.

\bibliographystyle{abbrv}
\bibliography{drift}  

\begin{thebibliography}{10}

\bibitem{globaltemp}
Global historical climatology network.
\newblock
  {https://www.ncdc.noaa.gov/data-access/land-based-station-data/land-based-datasets/global-historical-climatology-network-ghcn},
  2018.

\bibitem{Adams:aa}
R.~P. Adams and D.~J. MacKay.
\newblock Bayesian online changepoint detection.
\newblock {\em arXiv preprint arXiv:0710.3742}, 2007.

\bibitem{aminikhanghahi2017survey}
S.~Aminikhanghahi and D.~J. Cook.
\newblock A survey of methods for time series change point detection.
\newblock {\em Knowledge and information systems}, 51(2):339--367, 2017.

\bibitem{auger1989algorithms}
I.~E. Auger and C.~E. Lawrence.
\newblock Algorithms for the optimal identification of segment neighborhoods.
\newblock {\em Bulletin of mathematical biology}, 51(1):39--54, 1989.

\bibitem{bodenham2017continuous}
D.~A. Bodenham and N.~M. Adams.
\newblock Continuous monitoring for changepoints in data streams using adaptive
  estimation.
\newblock {\em Statistics and Computing}, 27(5):1257--1270, 2017.

\bibitem{disabato2021tiny}
S.~Disabato and M.~Roveri.
\newblock Tiny machine learning for concept drift.
\newblock {\em arXiv preprint arXiv:2107.14759}, 2021.

\bibitem{nile}
J.~Durbin and S.~J. Koopman.
\newblock {\em Time Series Analysis by State Space Methods}.
\newblock Oxford University Press, 2001.

\bibitem{gama2014survey}
J.~Gama, I.~{\v{Z}}liobait{\.e}, A.~Bifet, M.~Pechenizkiy, and A.~Bouchachia.
\newblock A survey on concept drift adaptation.
\newblock {\em ACM computing surveys (CSUR)}, 46(4):1--37, 2014.

\bibitem{hinder2020towards}
F.~Hinder, A.~Artelt, and B.~Hammer.
\newblock Towards non-parametric drift detection via dynamic adapting window
  independence drift detection (dawidd).
\newblock In {\em International Conference on Machine Learning}, pages
  4249--4259. PMLR, 2020.

\bibitem{ho2007detecting}
S.-S. Ho and H.~Wechsler.
\newblock Detecting changes in unlabeled data streams using martingale.
\newblock In {\em IJCAI}, pages 1912--1917, 2007.

\bibitem{ecp_paper}
N.~A. James and D.~S. Matteson.
\newblock {ecp}: An {R} package for nonparametric multiple change point
  analysis of multivariate data.
\newblock {\em Journal of Statistical Software}, 62(7):1--25, 2014.

\bibitem{ecp_package}
N.~A. James, W.~Zhang, and D.~S. Matteson.
\newblock {ecp}: An {R} package for nonparametric multiple change point
  analysis of multivariate data, 2019.
\newblock R package version 3.1.2.

\bibitem{kifer2004detecting}
D.~Kifer, S.~Ben-David, and J.~Gehrke.
\newblock Detecting change in data streams.
\newblock In {\em VLDB}, volume~4, pages 180--191. Toronto, Canada, 2004.

\bibitem{killick2012optimal}
R.~Killick, P.~Fearnhead, and I.~A. Eckley.
\newblock Optimal detection of changepoints with a linear computational cost.
\newblock {\em Journal of the American Statistical Association},
  107(500):1590--1598, 2012.

\bibitem{rcp}
R.~Killick, K.~Haynes, and I.~A. Eckley.
\newblock {\em {changepoint}: An {R} package for changepoint analysis}, 2016.
\newblock R package version 2.2.2.

\bibitem{kullback1951information}
S.~Kullback and R.~A. Leibler.
\newblock On information and sufficiency.
\newblock {\em The annals of mathematical statistics}, 22(1):79--86, 1951.

\bibitem{lorden1971}
G.~Lorden.
\newblock Procedures for reacting to a change in distribution.
\newblock {\em Ann. Math. Statist.}, 42(6):1897--1908, 12 1971.

\bibitem{Page1954}
E.~S. Page.
\newblock {Continuous inspection schemes}.
\newblock {\em Biometrika}, 41(1-2):100--115, 06 1954.

\bibitem{Page1955}
E.~S. Page.
\newblock {A test for a change in a parameter occurring at an unknown point}.
\newblock {\em Biometrika}, 42(3-4):523--527, 12 1955.

\bibitem{roberts2000control}
S.~Roberts.
\newblock Control chart tests based on geometric moving averages.
\newblock {\em Technometrics}, 42(1):97--101, 2000.

\bibitem{scott1974cluster}
A.~J. Scott and M.~Knott.
\newblock A cluster analysis method for grouping means in the analysis of
  variance.
\newblock {\em Biometrics}, pages 507--512, 1974.

\bibitem{Tartakovsky2006DetectionOI}
A.~Tartakovsky, B.~Rozovskii, R.~B. Blazek, and H.~Kim.
\newblock Detection of intrusions in information systems by sequential
  change-point methods.
\newblock {\em Statistical Methodology}, 3:252--293, 2006.

\bibitem{truong2020selective}
C.~Truong, L.~Oudre, and N.~Vayatis.
\newblock Selective review of offline change point detection methods.
\newblock {\em Signal Processing}, 167:107299, 2020.

\bibitem{9513865}
R.~Uppal, S.~Nagaraj, E.~van Leer, and D.~V. Anderson.
\newblock Non-parametric online changepoint detection algorithm.
\newblock In {\em 2021 IEEE Statistical Signal Processing Workshop (SSP)},
  pages 396--400, 2021.

\bibitem{wishart1928generalised}
J.~Wishart.
\newblock The generalised product moment distribution in samples from a normal
  multivariate population.
\newblock {\em Biometrika}, pages 32--52, 1928.

\bibitem{xie2019asynchronous}
L.~Xie, Y.~Xie, and G.~V. Moustakides.
\newblock Asynchronous multi-sensor change-point detection for seismic tremors.
\newblock In {\em 2019 IEEE International Symposium on Information Theory
  (ISIT)}, pages 787--791. IEEE, 2019.

\bibitem{rpackage}
Z.~Zanussi.
\newblock {\em {onlineCPD}: Detect Changepoints in Multivariate Time Series},
  2016.
\newblock R package version 1.0.

\end{thebibliography}

\end{document}